# Plurification in/of language technology
## The integration of 'culture' in next-generation AI

Gertraud Koch[1][0000-0002-2457-3335] and Fausto Giunchiglia[2][0000-0002-5903-6150]

[1] University of Hamburg, Edmund-Siemers-Allee 1 West, Germany
[2] University of Trento, Dipartimento Ingegneria e Scienza dell'Informazione
gertraud.koch@uni-hamburg.de

**Abstract.** The paper explores how 'culture' can be operationalised in Natural Language Processing (NLP) and what this reveals about the possibilities and limits of considering a plurality of cultural backgrounds in technological design. It proposes that cultural alignment cannot be achieved only by adding more examples of 'other cultures', rather it requires plural epistemologies: allowing multiple, locally grounded ways of knowing. To analyze how this plurality of knowing can be addressed in NLP, the paper uses a socio-technical model of language technology (LT) design, the five layers of technological activity model, for collecting and systematizing approaches to culture in NLP. The analysis shows that while NLP research has made progress toward more culturally sensitive systems, many approaches remain partial, addressing 'culture' primarily at the level of output or representation while leaving deeper questions of power, governance, and social context unresolved. The paper concludes that operationalising culture requires much more than technical adaptation; it suggests a reflexive and plural socio-technical approach that navigates potentials and limits of computational formalisation for accounting multiple linguistic and socio-cultural backgrounds.

## 1 Introduction

Language models and related language technologies should consider 'culture' because linguistic meaning is shaped not only by grammar and vocabulary, but also by socio-cultural norms, practices and common knowledge. In machine translation, a literal rendering may preserve propositional content while failing to convey culturally appropriate levels of expression or idiomatic meanings, particularly where honorifics, address forms, or culture-specific references require adaptation rather than direct transfer (House, 2015; Kenny, 2022). In dialogue systems, cultural situatedness is similarly essential and systems fail to model language's social factors adequately or ethically (Hovy and Spruit 2016, Hovy and Yang, 2021). For large language models (LLMs), socio-cultural backgrounds are both especially important and problematic because training data drawn from dominant internet discourse overrepresent majority perspectives and underrepresent minority voices, local knowledge, and culturally specific meanings (Blodgett et al., 2020; Bender et al., 2021). Low-resourced minority languages and languages distant from each other increase the complexity of machine translation rules (Mager et al., 2018; Nekoto et al., 2020) and meet an enormous variety of language practices like multilingualism, code switching and languaging.

The state of the art considering 'culture' in language technology is still exploratory. Current approaches tend to address 'culture' indirectly through adjacent constructs such as social factors of language, pragmatics, bias, and adaptation to low-resource or underrepresented communities (Hovy and Yang, 2021; Blodgett et al., 2020). In machine translation, culture is treated on a superficial level of honorifics, politeness, and context-sensitive meaning (House, 2015; Kenny, 2022). Similarly, in dialogue systems and LLMs, culture is often handled through prompt design, safety tuning, or post hoc moderation, yet these methods remain shallow because they do not manage to model the social norms and background knowledge that shape interpretation in the first place (Bender et al., 2021; Liu et al., 2024). The main difficulty is that culture is not a stable or uniformly measurable variable: it is dynamic, internally diverse, and deeply entangled with language variety, identity, power, and context. Modelling 'culture' is difficult because it includes but also goes beyond data collection, annotation, and evaluation, since culturally informed judgments often depend on specific expertise, cultural knowledge is distributed across populations and cannot be captured by generic benchmarks (Mager et al., 2018; Nekoto et al., 2020). The time dimension, that is, the fact culture evolves in time is maybe the most critical dimension, from both an anthropological and a technological perspective. There is an issue of how to track change in time as well as the never-ending high costs that this will require.

As a result, while there is growing recognition that 'culture' must be considered in language technology, methods for representing, learning, and evaluating culture-aware behaviour still remain a major open research problem. There are reasonable doubts that cultural alignment can be achieved only by adding more examples of 'other cultures' to the current NLP approaches. Rather it requires plural ontologies of the sociotechnical NLP systems for allowing multiple, locally grounded ways of knowing.

## 2 Where cultural diversity 'sits' in natural language processing: state of research

The discussion in natural language processing (NLP) about cultural diversity, multilingualism and variation is already wide-ranging and multi-faceted. However, culture is a concept too broad to study as a whole, which feeds back into NLP technology design. In NLP literature, the discussions about 'culture' overlap in several partly strands: (a) fairness, values and bias in NLP; (b) participatory and community-based NLP; and (c) socio-linguistically informed language technology. A foundational claim is that NLP systems are sociotechnical rather than purely technical, so design, deployment and use are inseparably entangled, and 'culture' is embedded in all elements of NLP systems. Datasets, annotation schemes, software and model architectures, and deployment contexts are all meaningful for implementing cultural diversity in a systems and shape how a system can become meaningful to users (Bender and Friedman 2018; Bender et al. 2021, Blodgett et al. 2020).

A more recent discussion is generating a growing body of work that addresses how cultural diversity and socio-cultural concerns, such as fairness, values and bias in NLP can be operationalized and evaluated (e.g. Mitchell et al. 2019, Raji et al. 2020, AlKhamissi et al. 2024). They introduce possibilities for considering cultural aspects in the design on language technology, often focusing on improving the performance of large language models (e.g. Choenni et al. 2024, Ren et al 2024). Other works emphasize the limits of LLMs in terms of modeling diversity. Giunchiglia points out that, while the existence of shared cross-lingual conceptualizations is obvious—otherwise no interlingual communication could ever be possible—the world's languages are, however, also extremely diverse on every level and captures this idea through the notions of *lexical gap* and *language specific words*, which are often missing for systematical reasons in LT (Giunchiglia, Batsuren and Freihat, 2018; Khishigsuren et al. 2022).

Participatory and community-based language technology are in the focus of a broader research area that develops ways to include the user needs and perspectives into technology design processes. Work on low-resource and indigenous-language NLP has helped shift attention from abstract notions of inclusion to concrete community participation, showing that local expertise improves both resource development and system relevance (e.g. Bird and Yaberbuk 2024, Mager et al. 2018, Nekoto et al. 2020). More broadly, language technology should support sustainable language development rather than only short-term model performance (Bird 2020).

The perspectives of socio-linguistic informed NLP show a manifold and complex interplay between language in everyday use and technology, raising issues such as representations avoiding stereotypes or fixed categories, accounting for internal diversity within language communities, developing benchmarks and evaluation protocols reflecting the variety of local norms, involving speakers, translators, educators, and community institutions from the beginning, avoiding exclusion and exploitation of language groups and users and studying how prompting, alignment, and retrieval of LLMs effect social and cultural life (eg Curry et al 2024, Priya et al. 2024, Lutgen et al. 2026).

This multi-facetedness of cultural dimensions in the NLP discussion is obvious, however not very systematically operationalized, which may be an effect of complexity and the dynamics of 'culture' as a concept and its range and variety in practice of everyday life. The integration of this plurality of socio-cultural backgrounds is at the core of considering 'culture' in NLP.

## 3 How to operationalize 'culture' in NLP

The literature review indicates that NLP is aware that 'culture' cannot be treated as a single, stable variable, nor can it be adequately captured through surface-level textual variation alone but demands theoretical grounded ideas for bringing technology and socio-cultural contexts together. Including 'culture' in NLP is not primarily a modelling task but a socio-technical problem, thus benefitting from a broader approach to the problem. An interdisciplinary perspective is the model of technological activity in language technology (Koch et al. 2024), which is the outcome of a collaboration of computer science, anthropology of technology and ethics in artificial intelligence aligns NLP with anthropology of technology on a high theoretical level of all involved disciplines. The model has been developed as a heuristic for studying and assessing cultural diversity in NLP systems in depth and detail.

According to this model, the analysis of socio-technical systems is organised around five layers of technological activity in NLP design processes (see Fig. 1). The first layer concerns the technological core, that is, the enabling software language technologies themselves. This includes model architectures, training procedures, decoding strategies, prompting methods, retrieval mechanisms, adapters and other computational procedures through which language is generated, transformed or interpreted. In relation to culture, the question is whether the technical core supports socio-culturally situated behaviour in a meaningful way or whether it merely produces surface-level stylistic variation, as for example Giunchiglia, Batsuren and Freihat (2018) have highlighted.

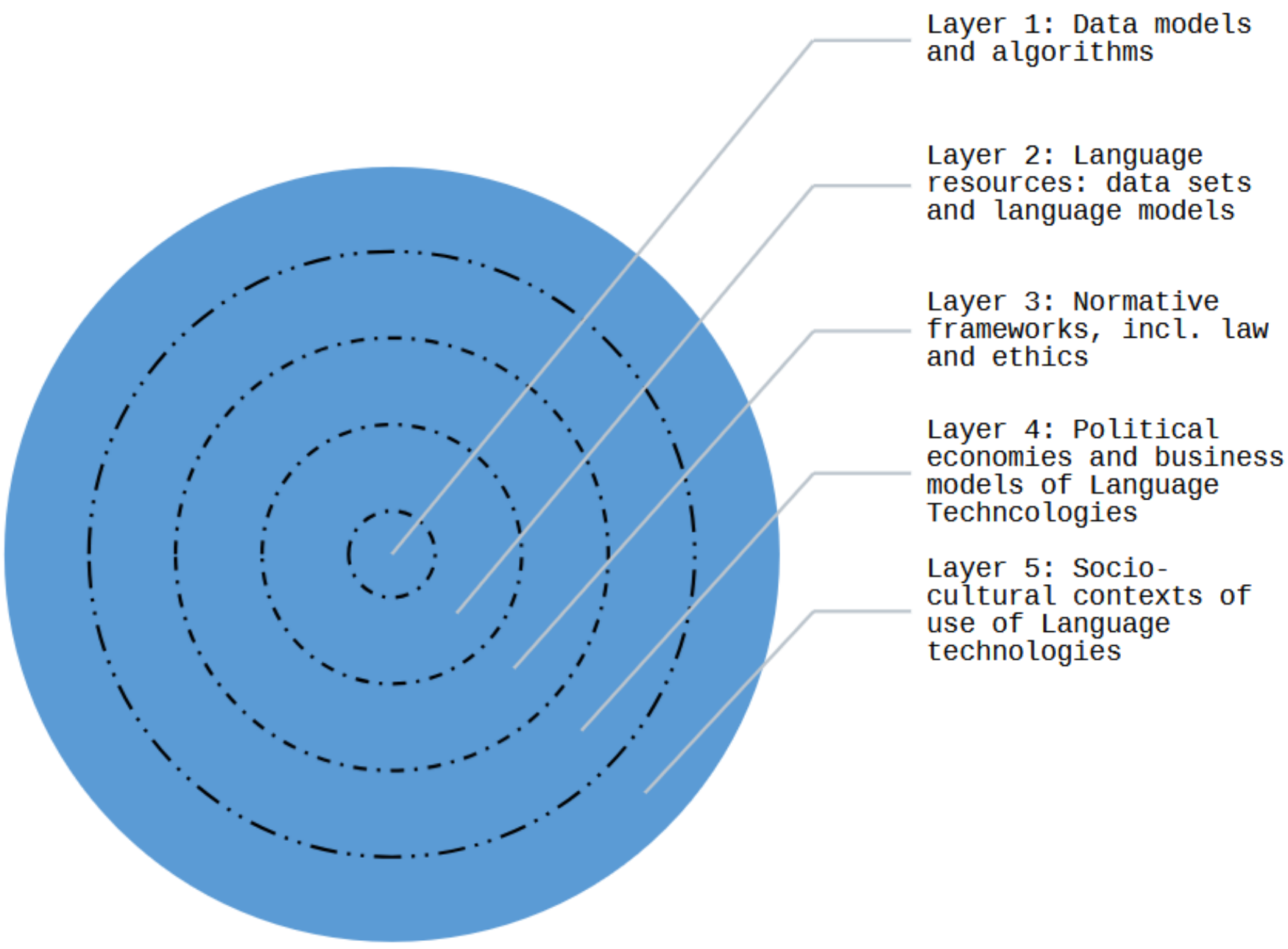

Figure 1 Layers of technological activity (Koch et al. 2024).

The second layer concerns the knowledge domain modelled in the system. This includes the corpora, linguistic resources, texts and symbolic categories from which the model learns. Here the focus is on representational questions: whose language, knowledge and cultural forms are treated as legitimate or normal? This is important because datasets are not neutral containers but constructed artefacts with social histories (Bender and Friedman, 2018), and language models trained on broad web data may reproduce dominant discourse while underrepresenting minority perspectives (Bender et al., 2021). The third layer concerns diverse knowledge ecologies, meaning the broader social and normative environments in which language technology operates. This includes local knowledge systems, myths, meanings, pragmatic conventions and the norms codified in legal and ethical frameworks. Hovy and Yang (2021) among others argue that NLP systems must model social factors such as speaker identity, audience, register and pragmatics if they are to interpret language reliably. The fourth layer concerns the political economies that make language technologies possible and maintain them. This includes data extraction practices, annotation labour, institutional funding, platform governance and power asymmetries. Participatory and community-based NLP has shown that language technologies can become extractive if communities are treated only as data sources rather than as co-designers or beneficiaries (Mager et al., 2018; Nekoto et al., 2020, Bird 2024). The fifth layer concerns the contexts of use, that is, the actual settings, communities and social situations in which technologies are deployed and interpreted. Here one question is whether a system is used and evaluated in ways that are locally meaningful and socially acceptable. Benchmarks and generic automatic metrics often fail to capture pragmatic appropriateness or local validity, making context-specific assessment necessary (Jiang et al., 2024). Another more crucial and complex dimension that goes beyond algorithmic modelling is how NLP aligns with communities' social reality and becomes meaningful for their futures (Nekoto et al. 2020, Bird and Yibarbak 2024, Bird, Aquino and Gumbala 2024).

Methodologically, the paper applies a qualitative, case-based approach as established in anthropology of technology (Coupaye 2022, Wahlberg 2022), because this allows for a focused conceptual discussion and it foregrounds the ways in which plurification may occur across all layers of technological design, ranging from technical modelling and corpus design to participatory governance and context-sensitive deployment. For each of the five layers, selected examples from NLP research are collected that demonstrate theoretical well-informed attempts to integrate cultural dimensions into the socio-technical NLP systems. The aim cannot be exhaustiveness, as the emergent field of integrating 'culture' in NLP systems is too vague and fuzzy still for being projected into a structured overview. Instead, an analytical exemplification will show how different layers of technological activity on language technology can be opened up to cultural concerns, and how these approaches may contribute to plurification of considered knowledges and ways of knowing in the socio-technical system.

The guiding questions for selecting the examples are whether the example broadens the language, knowledge or norms count in NLP; whether it challenges dominant or universalising assumptions in model design; whether it involves local expertise, participatory practices or culturally situated knowledges; and whether it intervenes only at the output layer or is an embedded feature in all layers of technological activity. This allows the chapter to distinguish between shallow forms of cultural adaptation, such as prompt-based style changes or politeness rewriting, and more substantial forms of plurification, such as participatory dataset creation, culturally situated knowledge, or governance structures that redistribute epistemic authority. It also makes it possible to link cultural integration to the wider concerns raised in the literature on bias, representation and social meaning in NLP (Blodgett et al., 2020; Bender et al., 2021; Hovy and Yang, 2021) and meaningful NLP systems for language groups and communities (Nekoto et al. 2020, Bird and Yibarbuk 2024). At the current state of the art, it is important to remain aware that the examples are often more or less singular interventions that explore possibilities to pluralize NLP where culture enters NLP design.

# 4 Plurification of knowledge in language technology

Culture awareness cannot be achieved by adding more examples and languages in LT; it requires more sophisticated approaches to plural epistemologies for including the multiple, locally grounded ways of knowing. This chapter presents selected cases from NLP research that demonstrate how plurality and socio-cultural diversity can be operationalised across the five layers of technological activity introduced in the previous chapter. Not aiming for a comprehensive survey of all relevant work, it offers an analytically structured discussion of culture theory informed approaches that consider that 'culture' is multi-faceted, dynamic, internally diverse, and entangled with identity, power, and local context. The examples are selected because they show how cultural complexity, visible in plurality of linguistic practices, social norms, local knowledge and community-specific values, enter NLP systems at different points in the technological activities. Taken together, they suggest that culture-aware NLP goes beyond a single modelling feature or an output-level adaptation, but emerges through interactions between algorithms, data, knowledge ecologies, political economies and contexts of use.

## 4.1 At the core: infrastructures, algorithms and data models

Contributions to plurifying technological infrastructure, algorithms, and model architectures aim to make language technology genuinely applicable across nations, countries under consideration of the specific infrastructural hardware conditions in the Global South (Lehuedé 2022, Lester et al. 2026). At the architectural level, plurification requires models that combine shared representations with language-specific flexibility, including modular, adapter-based, and knowledge-based approaches (Giunchiglia, Batsuren and Bella, 2017; Giunchiglia, Batsuren and Freihat, 2018). At the algorithmic level, it includes multilingual transfer, low-resource learning, and bias-aware evaluation methods that are sensitive to linguistic diversity. Contributions to this problem help shifting language technology toward a more inclusive and structurally plural ecosystem.

Fausto Giunchiglia's Universal Knowledge Core (UKC) (Giunchiglia, Batsuren and Bella, 2017; Giunchiglia, Batsuren and Freihat, 2018) can be understood as an important contribution to the plurification of language technology, because it offers a way to build multilingual systems that are not centered on a single dominant language. In the context of language technology, plurification refers to the effort to design tools and infrastructures that support multiple languages on more equal terms, rather than treating English as the default pivot language. The UKC advances this aim by proposing a language-neutral semantic layer: instead of organizing knowledge primarily around words in one language, it represents meaning at the level of concepts. Individual languages can then map their own lexical forms onto this shared conceptual structure (Giunchiglia, Batsuren and Bella, 2017; Giunchiglia, Batsuren and Freihat, 2018; Giunchiglia et al. 2023). This approach is significant because it enables many-to-many multilingual relations. Rather than translating or aligning everything through English, languages can interact through the Universal Knowledge Core. In practical terms, this can support multilingual search, cross-lingual information access, terminology alignment, and translation. At the same time, the UKC does not erase linguistic diversity: different languages may lexicalize concepts in different ways, and these differences remain visible at the level of language-specific mappings. In this sense, the UKC helps language technology become more inclusive and structurally decentered, since it reduces the dependence on a single global language while still allowing shared access to knowledge (Giunchiglia, Batsuren and Freihat 2018). From the perspective of plurification, the main value of Giunchiglia's work is therefore that it provides a conceptual and technical infrastructure for interoperability without linguistic homogenization. It makes it easier to imagine language technologies in which lesser-resourced and non-dominant languages can participate more equally, because they are connected through a common semantic framework rather than forced into an English-centered architecture (Giunchiglia, Batsuren and Freihat 2018). The UKC thus contributes to a more plural language-technological ecosystem in which multiple languages and knowledge systems can coexist and interact.

### 4.2 Language as knowledge: data sets and models

The knowledge domain modelled in the system, namely the corpora, language varieties and linguistic resources from which it learns, is central to socio-cultural diversity because it determines whose language becomes visible to the model and whose communicative practices are treated as normative. This is especially relevant for 'culture'-aware NLP because it shows that the data model cannot be treated as neutral and shows how socio-cultural diversity enters via the data model. Dialect, code-switching, multilingualism, and minority language corpora are key evidence (Bender and Friedman, 2018). Bender and Friedman (2018) make a foundational contribution by arguing that datasets are constructed artefacts with social histories. Their notion of data statements highlights the importance of documenting speaker population, language variety and collection context. Documentation and auditing frameworks also matter at this layer because they create accountability for how systems are built and deployed (Gebru et al. 2021). Bender and Friedman (2018) and Mitchell et al. (2019) show that documentation is essential for transparency, while Raji et al. (2020) demonstrate the need for auditing to identify harms.

Steven Bird's work has been influential in shaping the language data model that NLP systems learn from—particularly the corpora, language varieties, and linguistic resources that are collected, described, and made usable for computational modeling. His research ties NLP's language data needs to the practices of language documentation. Language documentation focuses on collecting and preserving evidence of communicative practices, often for languages and varieties that are underrepresented in mainstream NLP corpora. By treating documentation as a pathway for building structured linguistic evidence, Bird's work supports expanding the set of languages that can realistically be targeted for computational modeling, thereby improving linguistic inclusion in model training and evaluation (Simons and Bird 2003). A further contribution concerns the language data layer of LT: how linguistic data are encoded, curated, and reused. Bird has argued for formal and interoperable approaches to linguistic annotation and resource representation. Such approaches reduce "data lock-in" to specific tools and projects and increase the likelihood that corpora and annotations produced in one context can be transferred into others. This infrastructural dimension is important for diversity because it lowers barriers to building NLP systems beyond the small set of languages with existing, mature data ecosystems. Together, these strands of work—(a) connecting NLP to language documentation, and (b) promoting formal annotation frameworks and interoperable resource practices—help deciding whose language becomes computationally available. In that sense, Bird's scholarship contributes directly to the socio-cultural stakes of NLP by influencing the composition and portability of the linguistic knowledge domain that systems (Simons and Bird 2003).

Fausto Giunchiglia adds an important complementary dimension to this discussion: while Steven Bird's work primarily strengthens the availability, description, and interoperability of linguistic resources, Giunchiglia's work addresses the conceptual organization of knowledge that makes such resources useful across languages, domains, and cultures. In particular, Giunchiglia's research on ontology-based knowledge representation (Giunchiglia and Fumagalli, 2019; Fumagalli et al. 2020, Giunchiglia and Bagchi, 2024), context (Giunchiglia, 1993), and multilingual semantic interoperability (Khishigsuren et al., 2022, Giunchiglia et al. 2023) is highly relevant to the socio-cultural diversity of NLP. His work shows that language resources are not enough on their own: models also need structured ways to represent meaning, domain assumptions, and context in order to avoid collapsing diverse linguistic practices into a single normative interpretation. Giunchiglia has worked extensively on ontologies, i.e. formal conceptual structures for representing entities and relations. Ontologies can support multilingual and cross-domain alignment by providing a shared conceptual layer beneath different linguistic expressions. This is valuable for NLP because it enables models to connect semantically equivalent terms across languages without assuming one language variety is the default source of meaning. A major contribution of Giunchiglia's work is the idea that different knowledge systems—datasets, ontologies, terminologies, and language resources—can be aligned through formal methods. This matters for socio-cultural diversity because it offers a way to connect heterogeneous language communities and domain-specific practices without forcing them into a single uniform schema.

In effect, Giunchiglia's work supports semantic interoperability rather than mere data sharing. Giunchiglia's perspective is especially relevant where NLP must bridge multiple languages and conceptual systems, such as in multilingual retrieval, translation, linked data, and knowledge graphs. His work helps ensure that diversity is handled at the level of semantics and conceptualization, not only at the level of corpus collection (Giunchiglia and Fumagalli 2019; Giunchiglia et al., 2023).

### 4.3 Normative Frameworks: law, politics, ethics and everyday knowledge

After the release of ChatGPT normative frameworks for governing datafication and digitalisation gained specific attention and a new momentum. They are particularly relevant also for plurification in/of LT as they set guiding principles for AI development and deployment. In the wide field of policy and governance approaches most prominently the African Union in the *AI for Africa* framework published 2021 commits to principles of diversity-aware AI innovation, of society-centered management of AI and the principle of AI for the good of all, which is included in other frameworks more or less explicitly (Leslie and Perini 2024). With the publication of ChatGPT in November 2022 the normative frameworks of company driven AI development have entered the arena forcefully with only superficial interests in plurification but marketing- and profit-centered principles of AI development (Gebru et al 2023). This gives little space for acknowledging local and regional knowledge and the pragmatics and orientations of everyday life in different social realms as suggested in supra-national frameworks (UNESCO 2022, 2024). Normative frameworks, on the political and legal levels are always behind and of limited scope in an agile AI ecosystem (Wachter, Mittelstadt and Floridi, 2017). However, in the current situation the gap is big and referred to as AI governance crisis (Leslie, Ashurst et al., 2024). The normative approaches: AI for social good, responsible AI, participatory AI or trustworthy AI in this light appear as shallow technological fixes and repair attempts because they do not touch upon the basic problems of AI language technology resulting from marketing and profit-oriented LT approaches (Gebru et al. 2023, Bender and Hanna 2025).

Independent but also in the light of the current situation, the normative frameworks are crucial and challenging for the plurification of LT at the same time, as diverse frameworks overlap, intersect, interact, interdepend or contradict each other. It remains an open and empirical question if and in which ways the governance approaches may become effective. In practice, many dilemmas, discontingencies, and paradoxes have to be navigated; rarely there is a direct, ideal way into design and deployment. On the macro level supra-, inter- and national organisations in unison foster open data ecosystems as a pathway for plurification of language data (Council of Europe 2024, African Union 2024, 2024, 2025), as this seems to be the most adequate form in open, democratic societies. However, that troubles in practice well-founded demands for data ownership, justice and sovereignty, which remain a conundrum of these policies, as highlighted specifically in Global South perspectives (Fendji 2026) but is given elsewhere in similar ways. Shifts towards plurification in LT demands a substantial reorientation of design paradigms through acknowledging the many ways of knowing the world and of fixing problems beyond computational modelling (Majaca 2026), thus uncommits from the imaginary of a 'super-intelligence'. Overall, on the normative level there is a visible gap in plurification in/of language technology on the grounds but promising approaches in the African context (Fendji 2026, DIGITAfrica 2026) beyond the race of the Big Tech AI language technology approaches.

### 4.4 Economies beyond extractivism: business models, politial economies, and resourcing for plurification of language technology

The political economy of language technology, including the infrastructures, labour relations, institutional settings and power asymmetries, make NLP systems possible. Here, social and cultural diversity are important and cannot be understood solely as a technical issue; it is also shaped by who collects data, who annotates it, who defines the objectives of the system and who benefits from deployment. The literature on participatory and community-based NLP is particularly valuable here because it shows how language technologies can become extractive if communities are treated merely as sources of data rather than as co-designers or beneficiaries.

Nekoto et al. (2020) provide one of the strongest alternatives through participatory machine translation for African languages that indicates the relevance of symbolic and cultural capital in technology development and the potentials of a community-based resourcing for NLP. They show that effective system development may depend on local knowledge of orthographic conventions, lexical choice, dialectal variation, and culturally appropriate language use. Adapting NLP to local norms requires not only technical adjustments but also shared decision-making in data collection, annotation, and evaluation, so that the resulting systems are both linguistically valid and socially meaningful. Their approach foregrounds local expertise and community involvement, thereby redistributing some epistemic authority to speakers themselves and finding new ways of resourcing for NLP. However, data exploitation and souvereignity remain problems in this project as all data are open accessible for being exploited by everyone. Another African perspective (Fendji 2026) suggests a reorientation to address the data constraints by following three principles for building ecologically sustainable, culturally protective and digitally empowering data infrastructures: (a) infrastructure before annotation, (b) continuity before comprehensiveness and (c) sovereignty before openness.

From an Australian background, Bird (2020, 2022) similarly argues for sustainable language technology that supports long-term language development rather than short-term performance. This shifts the criterion of success from benchmark optimisation to community benefit and plurifies the economic value generated from data, which are contributed by speakers, annotators, translators, and communities. So far NLP systems are built on large-scale language resources that are often extracted from public discourse, digital platforms, or community speech with little recognition of the value created by those whose language use makes the systems possible. This raises questions of fairness, compensation, ownership, and benefit-sharing: who contributes the linguistic material, who transforms it into profitable systems, and who receives the gains? The analyses of Sambasivan et al. (2021) invite the conclusion that fresh and alternate business models for the plurification of AI are needed. They show that much of AI's value depends on data work that remains hidden and undervalued, including annotation, cleaning, translation, and contextual interpretation. This labour is usually performed by workers or communities whose contributions are essential but not proportionately recognised in the economic returns of the system. In terms of governance, the plurification of NLP demands forms of fair share, recognition, and benefit-sharing for data contributors, which requires stronger norms around consent, compensation, community ownership, and revenue-sharing models in NLP development. Plurification requires more than diverse input data but institutional arrangements that recognise labour, support community agency and balance power relations and exploitation of data.

### 4.5 Socio-cultural contexts and knowing the world differently

The socio-technical paradigm emphasizes the relations build through LT and AI across people, environments and social realms; it broadens the view towards other ways of knowing, such as in Indigenous or local knowledge systems with their specific ways of perceiving the world and framing problems (Bird, Aquino and Gumbala 2024) community computing approaches (Winschiers-Theophilus, Zaman and Stanley 2019) and decolonisation of design practices in technology (Smith et al 2024). Acknowledging the epistemic diversity beyond scientific approaches new potentials for solving problems emerge. In this sense intercultural communication across communities becomes a question of mutual learning rather than efficiently communicating technical solutions and also brings attention to those dimensions of local knowledge that are most relevant for problem solving beyond and with NLP modelling.

The contexts in which NLP technologies are used, interpreted, and judged determine whether outputs are socially intelligible, normatively acceptable, and practically useful. From the standpoint of plurification, the key claim is that cultural appropriateness is not a fixed property that can be read off from a model's training data or from generic benchmarks. Instead, it is situated: the same linguistic form may be experienced as respectful in one community and as ineffective, intrusive, or even disrespectful in another. Therefore, "culture-aware" NLP cannot be validated solely through technical performance; it must be assessed against the lived realities of speech communities and the local institutions that govern communicative norms. This situatedness is inseparable from questions of power and control over language technologies. Many NLP deployments reproduce extractive relations by treating communities as sources of linguistic data while leaving decision-making, infrastructure, and long-term benefits under centralized control (Bird 2024). Such extractive patterns matter in day-to-day use: systems may technically "work" while failing locally because they do not align with community-established relational world views. In plurifying terms, this means that appropriateness is not only a matter of "better language handling," but also of who gets to define what counts as appropriate and why (Bird 2024). Contexts of use must be treated as epistemic sites, not merely deployment environments. Speech communities maintain local knowledge about when, how, and with what consequences language is used.

When NLP systems ignore that knowledge, they tend to produce outputs that may be linguistically fluent but socially misaligned—e.g., by flattening culturally meaningful variation, mishandling community-specific pragmatic conventions, or failing to respect boundaries about what kinds of language content should be mediated by technology. (Bird, Aquino, and Gumbula 2024, Bird and Yibarbuk 2024) Plurification can also involve organizational and representational "third spaces" between global NLP practices and local communicative ecologies. In contexts where communities already have high-value linguistic resources (e.g., locally governed language archives, community translation expertise, or institutionally maintained language standards), the question is not simply how to transfer resources into NLP pipelines, but how to create design spaces in which local frameworks can structure model goals, evaluation, and interpretation. In such third spaces, community norms are not appended at the output layer; they influence how language technology is expected to function in practice—how it is queried, how it responds, and how its results are interpreted (Bird and Yibarbuk 2024). "Deployment context" therefore includes histories of linguistic marginalization, risks of surveillance, and the possibility of harm through misclassification or misrepresentation. These considerations affect whether a system is usable, trusted, and sustainable. In this light, culturally appropriate NLP requires participatory governance and long-term commitment—so that systems can be revised when community norms shift, when new disagreements emerge, and when local evaluation reveals forms of social failure that technical metrics cannot capture (Bird 2020). In sum, plurification at the level of contexts of use means: assessing social validity as locally situated, treating speech communities as epistemic and normative authorities, and resisting extractive deployment models that relocate control away from those whose language practices are being mediated.

### 4.6 Synthesis

Plurification of language technology cannot be achieved by adding more languages, more examples, or superficial adaptations to model outputs. Instead, it requires a broader transformation of the socio-technical and epistemic conditions under which NLP is built, evaluated, governed, and used—so that multiple, locally grounded ways of knowing can coexist without being reduced to a single dominant standard. Across the five layers of technological activity, 'culture' enters NLP as an interaction between technical choices, representational commitments, normative frameworks, political economies, and contexts of use. At the infrastructural level, plurification involves building systems that do not treat one language or one set of representational assumptions as default but favor relational NLP. This includes multilingual and modular approaches that enable cross-lingual interoperability without forcing homogenization. At the knowledge level (the data and knowledge domain modelled), plurification depends on recognizing that datasets and linguistic resources are socially situated artefacts with specific collection histories, speaker populations, and selection effects. Documentation and auditing are necessary to make these histories legible and to identify harms and blind spots. At the level of knowledge organization, semantic interoperability work—including conceptual or ontology-based approaches—can help connect heterogeneous linguistic practices without collapsing them into a single normative schema. However, plurification also requires confronting the governance and political-economic conditions that shape what language technologies become in practice. Normative frameworks determine which text or speech is treated as valid, who can contest system outputs, and what accountability mechanisms exist when social harm occurs. Political economy matters because language technologies rely on labour—annotation, translation, cleaning, contextual interpretation—that is often hidden and unevenly rewarded. Extractive infrastructures can reproduce asymmetries even when models are technically improved: communities may contribute linguistic value while decision-making and benefits remain external. Appropriateness is not established once a model is trained; it is realized (or fails) in situated interactions between systems and speech communities; this means that community members can interpret, trust, and effectively use outputs in locally meaningful ways. Taken together, operationalizing 'culture' is as inherently plural and reflexive along the layers of technological activity as 'culture is dynamic, internally diverse, and structured by power. LT should be socio-technical systems that remain responsive to plurality, disagreement, and context-specific validity. Plurification, in this sense, is not a final state but an ongoing design orientation: it redistributes epistemic authority toward speech communities, makes socio-political and -economic conditions visible, and preserves the possibility of multiple locally grounded ways of knowing within language technology.

# 5 Conclusion: 'culture', pluralization, and operationalisation

This paper has explored possibilities of plurification of/in LT to meet the multi-facetedness of socio-cultural contexts that are most crucial for language use and dynamics. The literature shows that 'culture' is already addressed in many parts of NLP, for example through dialect modelling, multilingual and code-mixed data, participatory data practices, pragmatics, documentation, and culture-sensitive evaluation. Yet the analysis also shows that these interventions are partial, mostly at the level of output or representation, while deeper questions of power, governance, and context remain unresolved. The five layers of technological activity from the pluriversal design perspective provided a useful heuristic for analysing where culture enters language technology: in the technological core, the knowledge domain modelled, diverse knowledge ecologies, political economies, and contexts of use. 'Culture'-aware NLP is therefore not mainly a technical problem and cannot be reduced to a single variable, a dataset property, or a matter of output style. It is a socio-technical question, which broadens the options for plurifications of LT design enormously. The paper showed that 'culture' and plurification are relevant in all layers of technological activity in language technology design. An adequate acknowledgement of cultural diversity thus demands mainstreaming of plurification through all layers of technological activity in LT systems design. It can be only partially formalised because of the dynamic, multi-facetedness nature of socio-cultural contexts. From this perspective, plurification is not a final state but a design orientation, facilitating the pluriversality of human civilization and knowledge.

**Disclosure of Interests.** The authors have no competing interests to declare that are relevant to the content of this article.